\documentclass[sigconf, screen, nonacm]{acmart}

\AtBeginDocument{%
\providecommand\BibTeX{{%
\normalfont B\kern-0.5em{\scshape i\kern-0.25em b}\kern-0.8em\TeX}}}

\usepackage[ruled,vlined]{algorithm2e}
\usepackage{amsmath}
\usepackage{multirow}
\usepackage{subfigure}

\begin{document}

\title{Proximally Optimal Predictive Control Algorithm for Path Tracking of Self-Driving Cars}

\author{Chinmay Samak}
\email{cv4703@srmist.edu.in}
\orcid{0000-0002-6455-6716}
\authornote{Both authors contributed equally to this research.}
\author{Tanmay Samak}
\authornotemark[1]
\email{tv4813@srmist.edu.in}
\orcid{0000-0002-9717-0764}
\author{Sivanathan Kandhasamy}
\email{sivanatk@srmist.edu.in}
\orcid{0000-0002-5302-030X}
\affiliation{%
	\institution{SRM Institute of Science and Technology}
	\city{Kattankulathur}
	\state{Tamil Nadu}
	\country{India}
	\postcode{603203}
}


\begin{abstract}\label{sec:Abstract}
	This work presents proximally optimal predictive control algorithm, which is essentially a model-based lateral controller for steered autonomous vehicles that selects an optimal steering command within the neighborhood of previous steering angle based on the predicted vehicle location. The proposed algorithm was formulated with an aim of overcoming the limitations associated with the existing control laws for autonomous steering -- namely PID, Pure-Pursuit and Stanley controllers. Particularly, our approach was aimed at bridging the gap between tracking efficiency and computational cost, thereby ensuring effective path tracking in real-time. The effectiveness of our approach was investigated through a series of dynamic simulation experiments pertaining to autonomous path tracking, employing an adaptive control law for longitudinal motion control of the vehicle. We measured the latency of the proposed algorithm in order to comment on its real-time factor and validated our approach by comparing it against the established control laws in terms of both crosstrack and heading errors recorded throughout the respective path tracking simulations.
\end{abstract}



\begin{teaserfigure}
	\includegraphics[width=\textwidth]{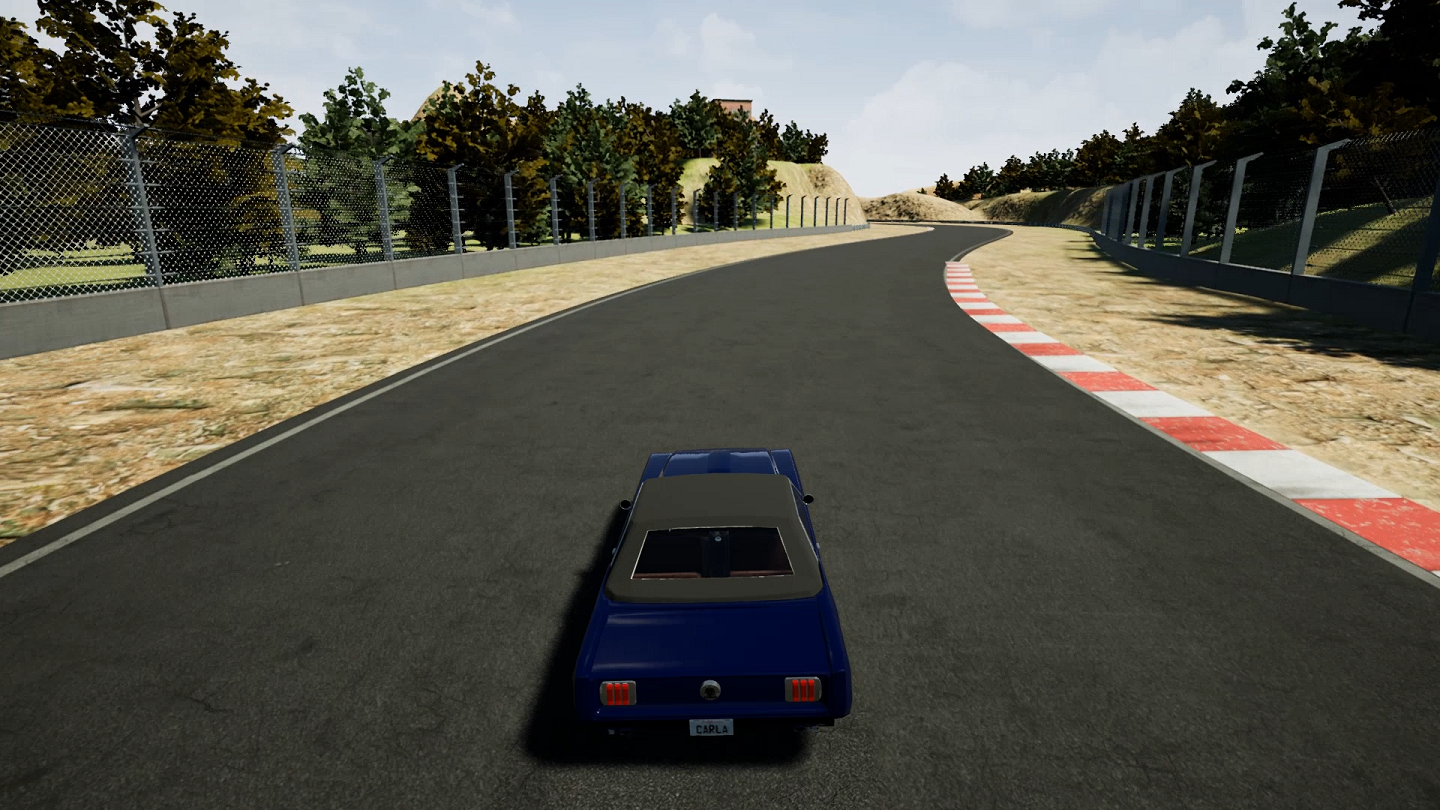}
	\caption{Autonomous path tracking using the proposed control algorithm in CARLA Simulator.}
	\Description{Screenshot of CARLA Simulator window running dynamic simulation of a vehicle being controlled autonomously for tracking a predefined path. The lateral motion is being controlled by the POP control algorithm, while the longitudinal motion is being controlled by an adaptive control law.}
	\label{fig:Teaser}
\end{teaserfigure}

\maketitle

\section{Introduction}\label{sec:Introduction}
Autonomous vehicles, or self-driving cars as they are popularly called, are complex engineering systems that are required to operate safely and efficiently in a rather broad operational design domain (ODD). It is a common practice, therefore, to divide the task of autonomous driving \cite{Yurtsever2020AutonomousDriving} into distinct \textit{perception}, \textit{planning} and \textit{control} modules:

\begin{itemize}
	\item \textbf{Perception:} This module employs a comprehensive sensor suite in order to measure the intrinsic and extrinsic parameters of the vehicle, which are then used for state-estimation and localization as well as object detection and tracking.
	\item \textbf{Planning:} This module plans a suitable behavior based on the perceived data and generates an optimal trajectory considering the global plan, behavior plan as well as hard and soft motion constraints.
	\item \textbf{Control:} This module controls the motion of the vehicle by actuating its final control elements (i.e. throttle, brake and steering), with an aim of accurately tracking the planned trajectory.
\end{itemize}

This work focuses on control of autonomous vehicles \cite{Samak2020ControlStrategies} by assuming a predefined trajectory generated offline. That being said, trajectory tracking requires complete two-dimensional motion control of the vehicle ensuring both, path tracking \cite{Snider2009AutomaticSteering} through lateral control (for spatial aspect) as well as velocity tracking through longitudinal control (for temporal aspect). This work employs an adaptive control law for longitudinal motion control of the vehicle and predominantly focuses on its lateral motion control for effective path tracking.

Generally, a path tracking algorithm may be implemented in numerous ways including simple model-free error-driven controllers (like Bang-Bang \cite{O'Brien2006Bang-Bang} or PID \cite{Marino2009NestedPID, Emirler2014RobustPID}), or kinematic/geometric controllers (like Pure-Pursuit \cite{Coulter1992Pure-Pursuit} or Stanley \cite{Hoffmann2007AutonomousAutomobile}), all the way up to model predictive controllers (MPC) \cite{Zanon2014ModelPredictiveControl, Babu2019ModelPredictiveControl, Obayashi2018Real-Time}. The simpler controller types offer real-time execution but often tend to be less accurate and robust in terms of tracking efficiency, while the complex ones provide better tracking at the cost of higher computational overhead. Thus, there is a need to formulate an efficient path tracking algorithm offering robustness and accuracy in real-time.

\section{Background}\label{sec:Background}

This section briefly describes three of the established path tracking controllers for autonomous vehicles, viz. PID, Pure-Pursuit and Stanley controllers. Their respective limitations are also discussed in this section.

\subsection{PID Controller}

PID controller makes use of proportional, integral and derivative terms of the error variable (along with respective tunable gains, viz. $k_P$, $k_I$ and $k_D$) to better control the system under consideration. It is a rather versatile model-free error-driven controller employed for controlling a wide range of engineering systems.

\begin{equation}
\delta_{t+1}=k_P*{^{\chi}e}_t+k_I*\sum_{i=t_0}^{t}{^{\chi}e}_i+k_D*\left[\frac{{^{\chi}e}_t-{^{\chi}e}_{t-1}}{\Delta t}\right]
\label{eqn:PID}
\end{equation}

In the context of lateral control of autonomous vehicles, a PID controller usually operates on the crosstrack error ${^{\chi}e}$ as defined in Equation \ref{eqn:PID}. However, its adoption for the said task is largely limited due to the fact that its tracking efficiency is severely affected beyond the operating point for which it was designed/tuned.

\subsection{Pure-Pursuit Controller}

Pure-Pursuit controller is essentially a geometric path tracking algorithm. It can be thought of as a proportional control law for steering that acts on the instantaneous crosstrack error ${^{\chi}e}$ of the vehicle. An adaptive Pure-Pursuit control law with velocity-based lookahead adaptation is given by Equation \ref{eqn:Pure_Pursuit}.

\begin{equation}
\delta=tan^{-1}\left(\frac{2*L*sin(\alpha)}{k_v*v}\right);~\delta \in \left[-\delta_{max},\delta_{max}\right]
\label{eqn:Pure_Pursuit}
\end{equation}
where, $L$ is wheelbase of the vehicle, $\alpha$ is the angle between heading vector of the vehicle and the line joining vehicle coordinate frame origin (located at the center of rear axle) to the lookahead point, and $k_v$ is a proportional coefficient to reduce the steering command at higher velocities $v$. The final control command is clipped within the upper and lower steering limits, $-\delta_{max}$ and $\delta_{max}$ respectively.

An evident limitation of this controller is that it does not act on the heading error. Another significant limitation, which may not be apparent enough, arises due to the lookahead-tracking nature of this algorithm -- there may arise event(s) wherein a lookahead point may not be available (e.g. towards the end of a trajectory), causing the controller to generate erratic actions.

\subsection{Stanley Controller}

Stanley control law was formulated by Dr. Gabe Hoffman of Stanford University for their autonomous vehicle ``Stanley'' that won the DARPA Grand Challenge (2005). Unlike Pure-Pursuit controller, which only acts on the crosstrack error term, Stanley controller generates steering command by operating on both crosstrack $({^{\chi}e})$ as well as heading $({^{\psi}e})$errors.

Considering front axle of the vehicle as the reference point, Stanley control law can be expressed using Equation \ref{eqn:Stanley}.

\begin{equation}
\delta={^{\psi}e}+tan^{-1}\left(\frac{k_\chi*{^{\chi}e}}{k_s+k_v*v}\right);~\delta \in \left[-\delta_{max},\delta_{max}\right]
\label{eqn:Stanley}
\end{equation}
where, $k_\chi$ is a proportional coefficient acting on the crosstrack error term, $k_s$ is a softening coefficient that avoids numerical instability arising due to extremely low vehicle velocities (in the near-neighborhood of zero), and $k_v$ is another proportional coefficient that reduces the steering command at higher velocities $v$. As with Pure-Pursuit controller, the final control command is bounded within the upper and lower steering limits, $-\delta_{max}$ and $\delta_{max}$ respectively.

It is to be noted that Stanley controller uses the closest waypoint on the trajectory to compute crosstrack error. Additionally, it acts directly on the heading error. Both of these facts can make it quite aggressive at low speeds (if tuned for high speeds) or sluggish at high speeds (if tuned for low speeds).

\section{Controller Design}\label{sec:Controller_Design}

As discussed earlier, path tracking requires both lateral as well as longitudinal motion control of the vehicle. This section explains the controller design adopted for this work.

\subsection{Lateral Controller}

This work proposes proximally optimal predictive (POP) controller, a lateral controller that ensures tight tracking in real-time. The POP controller (refer Algorithm \ref{alg:POP_Controller}) is formulated with an aim of bridging the gap between traditional and optimal controllers by defining a light-weight optimization loop that selects the best possible steering command within the neighborhood of previous steering angle in order to minimize the distance between predicted vehicle location and lookahead point on reference path. This theoretically gurantees path tracking performance, practical validation of which is elucidated in Section \ref{sec:Results}.

\begin{algorithm}[]
	\SetAlgoLined
	\caption{Proximally optimal predictive controller}
	\label{alg:POP_Controller}
	\KwIn{Vehicle states $\left\{x,y,\theta,v\right\}\in\mathbb{R}^4$, waypoints $\left\{w\right\}\in\mathbb{R}^N$, previous steering command $\delta_{prev}$, list of steering commands $\delta_{list}$ within given neighborhood $\delta_{nbd}$ i.e. $\delta_{list}=\left\{-\delta_{nbd},\cdots,0,\cdots,\delta_{nbd}\right\}\subset \left\{-\delta_{max},\delta_{max}\right\}$, velocity constant $k_v$, minimum lookahead distance $ld_{min}$}
	\KwOut{Vehicle steering command $\delta$}
	\nl $\delta_{list}+=\delta_{prev}$ \tcp*{$\delta_{list}$ in neighborhood of $\delta_{prev}$}
	\nl $ld=ld_{min}+k_v*v$ \tcp*{lookahead dist}
	\nl $lp_i=$ \textbf{get\_lookahead\_point\_index(}$x, y, \left\{w\right\}, ld$\textbf{)}\;
	\nl $lp=$ \textbf{get\_lookahead\_point(}$\left\{w\right\}, lp_i$\textbf{)}\;
	\nl $\left\|d_{lp}^{V_{pred}}\right\|_{2}^{min}\leftarrow\infty$ \tcp*{min dist to lp from vehicle}
	\nl $\delta=\delta_{prev}$ \tcp*{sanity check}
	\nl \For{k $\leftarrow$ 1 to \textbf{\texttt{len}(}$\delta_{list}$\textbf{)}}{
		\nl \tcc{predict vehicle location in future}
		\nl $V_{pred}=$ \textbf{get\_predicted\_location(}$x,y,\theta,v,\delta_{list}[k]$\textbf{)}\;
		\nl \tcc{dist b/w predicted vehicle location \& lp}
		\nl $\left\|d_{lp}^{V_{pred}}\right\|_{2}=$ \textbf{get\_distance(}$V_{pred},lp$\textbf{)}\;
		\nl \tcc{minimize dist from lp}
		\nl	\If{$\left\|d_{lp}^{V_{pred}}\right\|_{2}<\left\|d_{lp}^{V_{pred}}\right\|_{2}^{min}$}{
			\nl	$\delta=\delta_{list}[k]$\;
			\nl $\left\|d_{lp}^{V_{pred}}\right\|_{2}^{min}=\left\|d_{lp}^{V_{pred}}\right\|_{2}$\;
		}
	}
	\nl $\delta_{prev}=\delta$ \tcp*{update $\delta_{prev}$}
	\Return{$\delta$}\;
\end{algorithm}

The \textbf{\texttt{get\_distance}} function computes Euclidean distance between two points parsed as a pair of X and Y coordinates, $(x_1, y_1)$ and $(x_2, y_2)$, as described in Equation \ref{eqn:Get_Distance}.

\begin{equation}
\left\| d \right\|_2 = \sqrt{(x_2-x_1)^2 + (y_2-y_1)^2}
\label{eqn:Get_Distance}
\end{equation}

The \textbf{\texttt{get\_lookahead\_point\_index}} function loops through the list of available waypoints $\left\{w\right\}$ and iteratively computes the distance of each waypoint from the ego vehicle using \texttt{get\_distance} function. If the distance to a particular waypoint is almost equal to the lookahead distance $ld$, it returns the index $lp_i$ of that particular waypoint. Equation \ref{eqn:Get_LP_Index} describes the criteria for selecting a particular waypoint index. This work assumes a distance approximation threshold of $\epsilon$ = 10e-3 m.

\begin{align}
lp_i = 
\begin{cases}
i & \text{if~} \left\|d_{w_i}^{V}\right\|_{2} \leq \epsilon ~ \exists ~ w_i \in \{w\} \\
\texttt{len}(w)-1 & \mathrm{otherwise} \\
\end{cases}
\label{eqn:Get_LP_Index}
\end{align}

The \textbf{\texttt{get\_lookahead\_point}} function returns the actual waypoint object corresponding to a particular index.

\begin{figure}
	\includegraphics[width=0.87\linewidth]{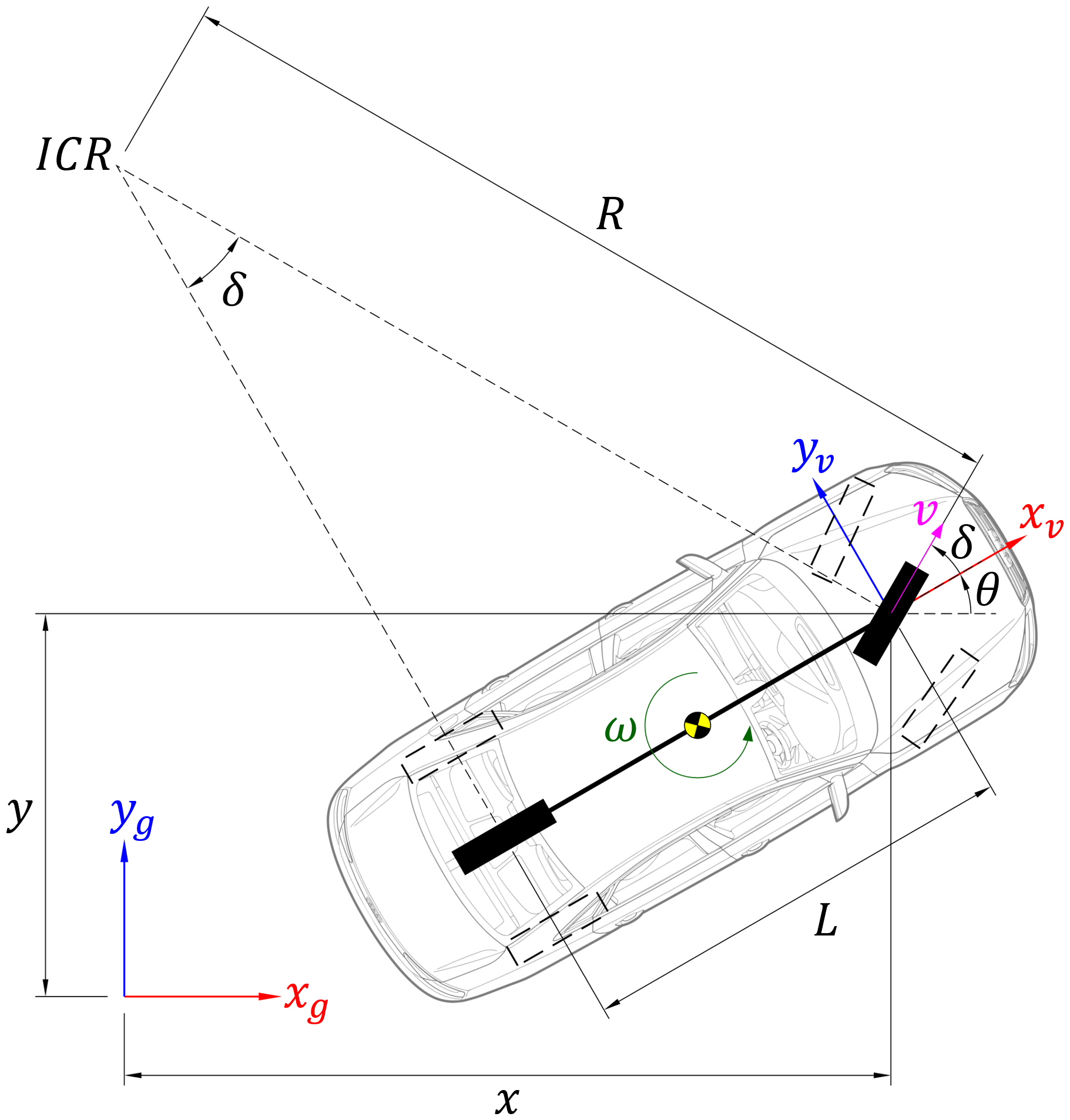}
	\caption{Vehicle kinematics.}
	\Description{Illustration of vehicle kinematics with reference to the kinematic bicycle model.}
	\label{fig:Vehicle_Kinematics}
\end{figure}

The \textbf{\texttt{get\_predicted\_location}} function returns the vehicle location $(x_{t+1}, y_{t+1})$ predicted ahead of time interval $\Delta t$ based on its current state and control input (steering angle). This work defines the vehicle's state transition equation (Equation \ref{eqn:Get_Predicted_Location}) based on the kinematic bicycle model (Figure \ref{fig:Vehicle_Kinematics}), which achieves sufficient accuracy; nevertheless, detailed dynamic models can be used for even better prediction (usually at the cost of a higher computational overhead).

\begin{equation}
\begin{bmatrix}x_{t+1}\\ y_{t+1}\\ \theta_{t+1}\\ v_{t+1}\\ \omega_{t+1}\end{bmatrix} =
\begin{bmatrix}1 & 0 & 0 & \cos\phi_t\cdot\Delta t & 0\\ 0 & 1 & 0 & \sin\phi_t\cdot\Delta t & 0\\ 0 & 0 & 1 & 0 & \Delta t\\ 0 & 0 & 0 & 1 & 0\\ 0 & 0 & 0 & 0 & 1\end{bmatrix}\times
\begin{bmatrix}x_t\\ y_t\\ \theta_t\\ v_t\\ \omega_t\end{bmatrix}+
\begin{bmatrix}0 & 0\\ 0 & 0\\ 0 & 0\\ 1 & 0\\ 0 & 1\end{bmatrix}\times
\begin{bmatrix}u_\tau\\ u_\delta\end{bmatrix}
\label{eqn:Get_Predicted_Location}
\end{equation}
where, $\phi$ is the sum of $\theta$ and $\delta$, which respectively represent the yaw (orientation) and steering angle of the vehicle (i.e. $\phi = \theta+\delta$), $[x, y]$ represents the location coordinates (position) of the vehicle, $v$ and $\omega$ respectively represent the linear and angular velocities of the vehicle, $u_\tau$ and $u_\delta$ respectively represent the longitudinal (throttle/brake) and lateral (steering) control inputs to the vehicle, and $\Delta t$ represents time interval. Note that $t$ in the subscript denotes value of a particular variable at current time instant, while $t+1$ denotes its value at the next time instant.

The POP controller can be tuned based on vehicle dynamics and driving behavior. Other tuning criteria include the required degree of optimality and controller loop rate. Following are the tunable parameters for POP controller:
\begin{itemize}
	\item \textbf{Velocity constant:} The velocity constant $k_v$ alters the lookahead distance in direct proportion to the vehicle velocity. This helps the controller adapt to both lower and higher speeds.
	\item \textbf{Optimization range:} The optimization range is controlled by $\delta_{nbd}$ parameter. For a constant value of optimization resolution, larger range implies wider limits (usually recommended for rigorous driving) whereas lower range implies higher resolution (meaning higher degree of optimality).
	\item \textbf{Optimization resolution:} The optimization resolution is controlled by the size of $\delta_{list}$. A larger size implies higher resolution, which results in higher degree of optimality (at the cost of higher computational overhead). On the other hand, smaller size implies lower resolution, resulting in faster execution (at the cost of sub-optimal solutions).
\end{itemize}

\subsection{Longitudinal Controller}

A longitudinal controller can be either implemented independently or coupled with the lateral controller. The former allows direct tracking of a velocity setpoint (constant or time-varying) while the later offers more stable and safe control of the vehicle. Since our problem statement was to address path tracking (without explicit focus on velocity tracking), we adopted a modified form of longitudinal controller described in \cite{Samak2021RobustBehavioralCloning} in order to generate throttle/brake commands for the vehicle. The control law is formulated as Equation \ref{eqn:Longitudinal_Controller}.

\begin{equation}
\tau = k_\tau+\left[\frac{v_{lim}-v}{v_{lim}}-\frac{\left|\delta\right|}{\delta_{lim}}\right]*(1-k_\tau)
\label{eqn:Longitudinal_Controller}
\end{equation}
where, $\tau$ is the longitudinal control command (positive values imply throttle while negative ones imply brake), $\delta$ is the commanded steering angle, $\delta_{lim}$ is the steering limit of the vehicle, $v$ is the vehicle velocity, $v_{lim}$ is the prescribed speed limit of the vehicle, and $k_\tau$ is the cutoff constant (it controls the offset and aggressiveness of the longitudinal control command).

\section{Implementation Details}\label{sec:Implementation_Details}

In order to test the proposed control algorithm and validate its efficiency against the established counterparts, we implemented PID, Pure-Pursuit, Stanley and POP control algorithms\footnote{Code: \url{https://github.com/Tinker-Twins/Self_Driving_Car_Trajectory_Tracking}} (controller parameters are summarized in Table \ref{tab:Lateral_Controller_Parameters}) in Python 3.6.5 and ran dynamic simulation tests for each of them using CARLA Simulator 0.8.4 \cite{Dosovitskiy2017CARLA}. All the tests were carried out on a laptop PC incorporating Intel i7-8750H CPU, NVIDIA RTX 2070 GPU and 16 GB RAM. The simulation timestep was fixed at 50 ms.

\begin{table}[h]
	\caption{Lateral controller parameters}
	\label{tab:Lateral_Controller_Parameters}
	\begin{tabular}{lll}
		\toprule
		\textbf{Controller}      & \textbf{Parameter}                                         & \textbf{Value} \\
		\midrule
		\multirow{4}{*}{PID}     & Proportional gain, $k_P$                                   & 0.25           \\
		& Integral gain, $k_I$                                       & 0.01           \\
		& Derivative gain, $k_D$                                     & 0.2            \\
		& Buffer length\footnotemark                                 & 500            \\
		\midrule
		Pure-Pursuit             & Velocity constant, $k_v$                                   & 0.9            \\
		\midrule
		\multirow{3}{*}{Stanley} & Crosstrack error constant, $k_\chi$                        & 1.5            \\
		& Velocity constant, $k_v$                                   & 1.3            \\
		& Softening constant, $k_s$                                  & 1e-5           \\
		\midrule
		\multirow{3}{*}{POP}     & Velocity constant, $k_v$                                   & 0.2            \\
		& Optimization range, $\left|\delta_{nbd}\right|$ & 3             			  \\
		& Optimization resolution                                    & 21             \\
		\bottomrule
	\end{tabular}
\end{table}
\footnotetext{The problem of integral windup was handled by defining a dynamic first-in-first-out (FIFO) buffer to accumulate only the most recent error values.}

The path to be followed by the vehicle was generically defined by 1274 waypoints (with an inter-waypoint resolution of $\approx$ 1 m), which was too sparse for effective tracking. The waypoints in near-neighborhood of the vehicle were therefore interpolated online to form a continuous path, which was then re-discretized at a higher resolution (0.01 m).

\begin{table}[h]
	\caption{Longitudinal controller parameters}
	\label{tab:Longitudinal_Controller_Parameters}
	\begin{tabular}{ll}
		\toprule
		\textbf{Parameter}                                         & \textbf{Value} \\
		\midrule
		Cutoff constant, $k_\tau$                                  & 0.5            \\
		Steering limit, $\delta_{lim}$                             & 1.22 rad       \\
		Speed limit, $v_{lim}$                                     & 69.44 m/s      \\
		\bottomrule
	\end{tabular}
\end{table}

In order to maintain homogeneity throughout all the experiments, the environment conditions and vehicle model were kept constant. Additionally, the longitudinal controller parameters (refer Table \ref{tab:Longitudinal_Controller_Parameters}) were kept identical for all the implementations to avoid its varied effect on the lateral controller in terms of path tracking.

\section{Results}\label{sec:Results}

\begin{figure}[t]
	\centering
	\subfigure[]{\includegraphics[width=0.49\linewidth]{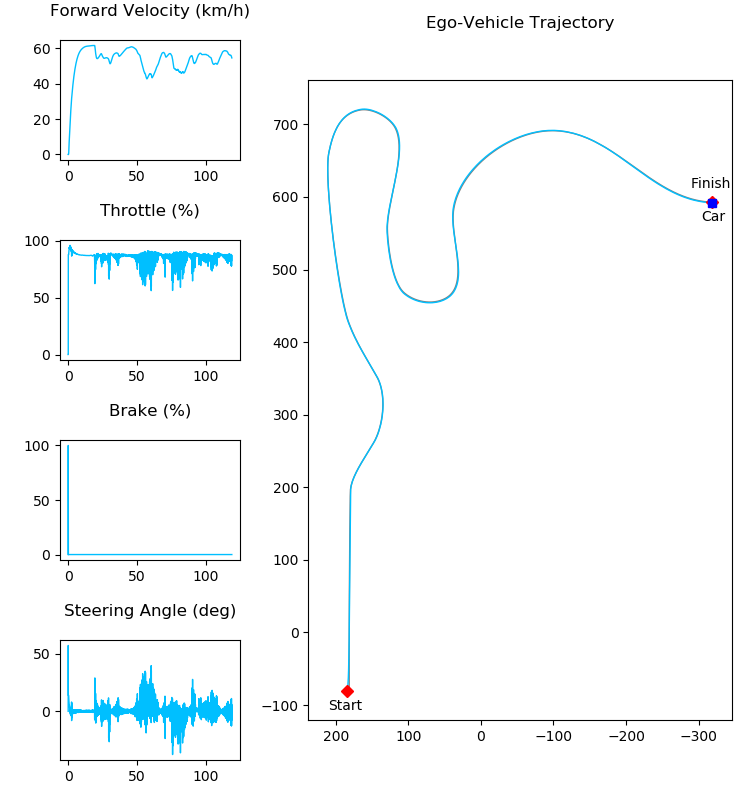}}
	\subfigure[]{\includegraphics[width=0.49\linewidth]{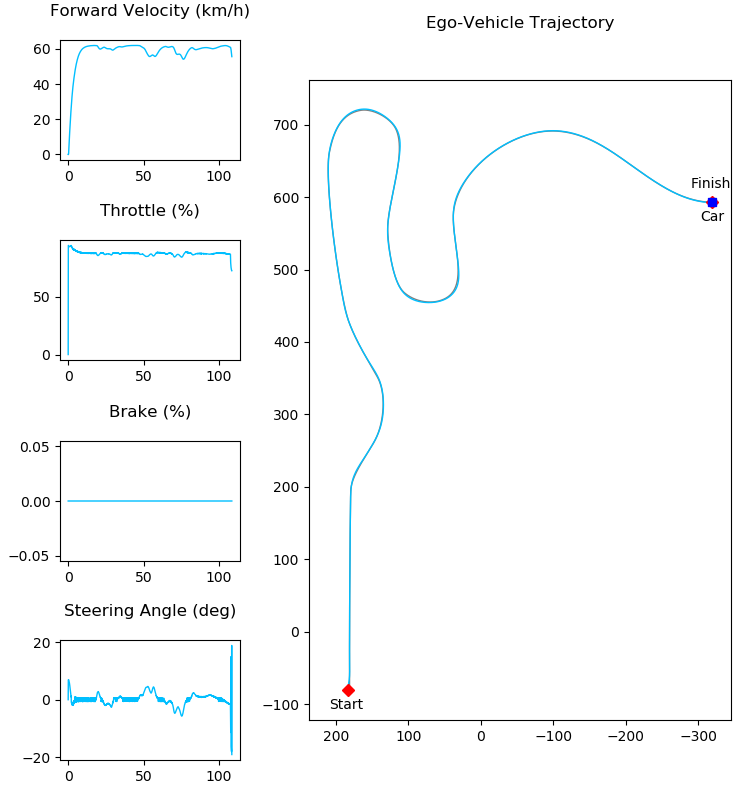}}
	\subfigure[]{\includegraphics[width=0.49\linewidth]{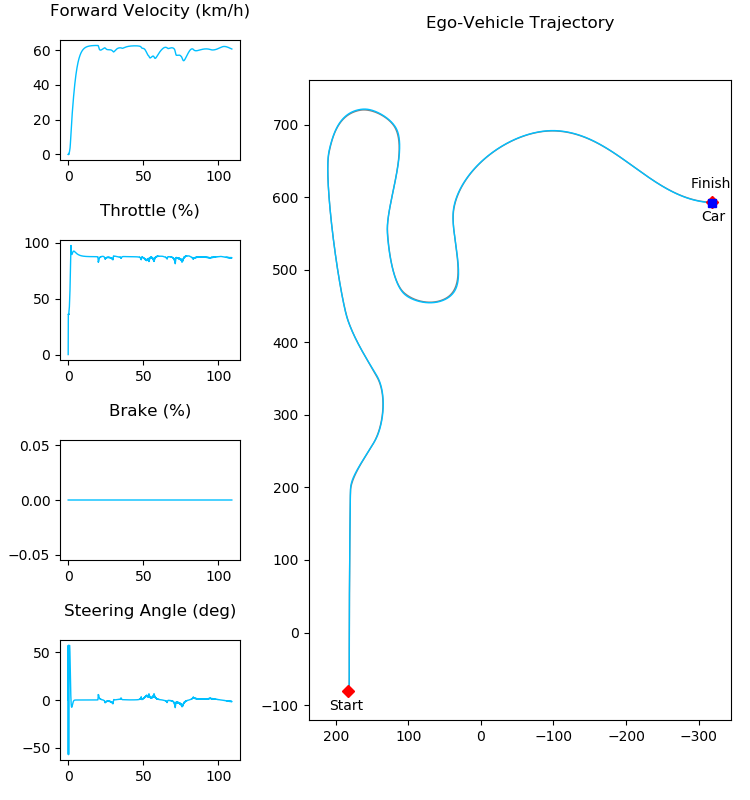}}
	\subfigure[]{\includegraphics[width=0.49\linewidth]{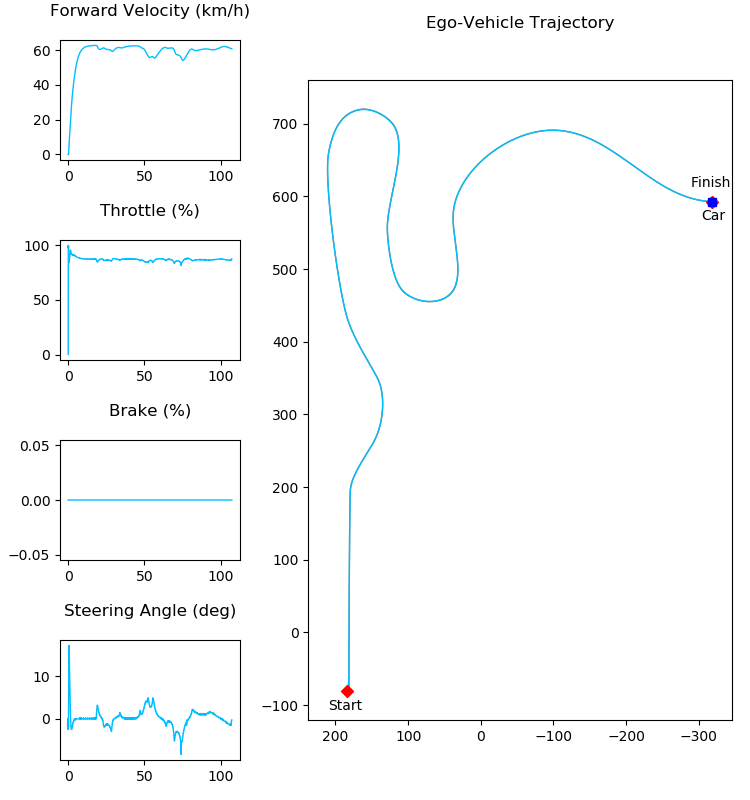}}
	\caption{Tracking performance\protect\footnotemark of (a) PID, (b) Pure-Pursuit, (c) Stanley, and (d) POP controllers.}
	\Description{Path tracking performance of self-driving car using (a) PID, (b) Pure-Pursuit, (c) Stanley, and (d) POP controllers as per the dynamic simulation tests in CARLA Simulator.}
	\label{fig:Results}
\end{figure}

Figure \ref{fig:Results} illustrates the path tracking performance of the simulated self-driving car using established controllers as well as the POP control algorithm proposed in this work. Table \ref{tab:Tracking_Metrics} supplements these results by providing mean absolute metrics for both crosstrack and heading errors.

The inherent shortcomings of existing controllers, as discussed in Section \ref{sec:Background}, are evident from the respective steering command plots. Particularly, the PID controller struggled throughout the course, right from the start where it generated steering commands of over 50°, followed by high-frequency switching due to the dynamic variations beyond the operating point for which it was tuned. The Pure-Pursuit controller managed to control the vehicle pretty well throughout the course; however, towards the end, it generated erratic steering commands of about 20° in either direction, owing to the loss of lookahead point (in the actual experiments the lookahead point was never lost, the trajectory preprocessor returned the last waypoint as lookahead point towards the end of path, but the controller could not adapt to such a close lookahead distance due to the high velocity). Stanley controller exhibited promising results throughout the track, except during the start, where it aggressively generated steering commands of over 50° in either direction, owing to the high initial offset between the vehicle spawn point and the first waypoint on the defined path. The POP controller, overcame most of these limitations, if not all, and controlled the vehicle quite accurately, yet smoothly throughout the course; not to mention its tracking efficiency was the best compared with others.

\footnotetext{Video: \url{https://youtu.be/nPjSj4hp6N4}}

\begin{table}[t]
	\caption{Tracking metrics}
	\label{tab:Tracking_Metrics}
	\begin{tabular}{lcc}
		\toprule
		\textbf{Controller} & \textbf{Crosstrack Error (m)} & \textbf{Heading Error (rad)}	\\
		\midrule
		PID          		& 0.4958               			& 0.0121              			\\
		Pure-Pursuit 		& 0.3662               			& 0.0219              			\\
		Stanley      		& 0.3383               			& 0.0141              			\\
		POP          		& 0.1761               			& 0.0079             			\\
		\bottomrule
	\end{tabular}
\end{table}

Running parallel to the CARLA Simulator, a Python implementation of the POP controller exhibited an average latency of 3.239 ms, which is a pretty good value, especially considering that it involved trajectory pre-processing, model-based prediction and proximal optimization of the control command.  This translates to a control loop rate of little over 300 Hz, which can be very well considered as ``real-time'' performance, and is bound to improve several folds through the use of a compiled language and/or distributed computing.

\section{Conclusion}\label{sec:Conclusion}

In this work, we presented proximally optimal predictive controller for lateral motion control of steered non-holonomic vehicles in order to achieve efficient path tracking. We also presented an adaptive coupled control law for longitudinal control of the vehicle. Finally, a comparative analysis was presented so as to validate our algorithm against the established controllers employed for the said application.

In terms of path tracking, the POP controller proved to be most efficient, followed by Stanley, Pure-Pursuit and PID controllers. The entire loop of the proposed control algorithm (including trajectory pre-processing, model-based prediction and proximal optimization of the control command), while running alongside the CARLA Simulator, required about 3.239 ms ($\ge$300 Hz loop rate), thereby assuring a real-time implementation even with an interpreted language (Python).

%

\bibliographystyle{ACM-Reference-Format}
\bibliography{References}


\end{document}